\documentclass[10pt,twocolumn,letterpaper]{article}

\usepackage[pagenumbers]{cvpr} 

\usepackage[dvipsnames]{xcolor}

\definecolor{cvprblue}{rgb}{0.21,0.49,0.74}
\usepackage[pagebackref,breaklinks,colorlinks,citecolor=cvprblue]{hyperref}

\usepackage[linesnumbered,ruled,vlined,algo2e]{algorithm2e}
\usepackage{algpseudocode}
\usepackage[accsupp]{axessibility} 

\def\paperID{16034} 
\def\confName{CVPR}
\def\confYear{2024}

\title{\textsc{InitNO}: Boosting Text-to-Image Diffusion Models via Initial Noise Optimization}

\author{Xiefan Guo$^{1,2}$\quad Jinlin Liu\quad Miaomiao Cui\quad Jiankai Li$^{1,2}$\quad Hongyu Yang$^{3,4}$\quad Di Huang$^{1,2}$\thanks{Corresponding author.}\\[5pt]
$^{1}$State Key Laboratory of Software Development Environment, Beihang University, Beijing, China\\
$^{2}$School of Computer Science and Engineering, Beihang University, Beijing, China\\
$^{3}$Institute of Artificial Intelligence, Beihang University, Beijing, China\\
$^{4}$Shanghai Artificial Intelligence Laboratory, Shanghai, China\\
{\tt\small \{xfguo,lijiankai,hongyuyang,dhuang\}@buaa.edu.cn}
}

\begin{document}

\maketitle

\begin{abstract}
Recent strides in the development of diffusion models, exemplified by advancements such as Stable Diffusion, have underscored their remarkable prowess in generating visually compelling images. However, the imperative of achieving a seamless alignment between the generated image and the provided prompt persists as a formidable challenge. This paper traces the root of these difficulties to invalid initial noise, and proposes a solution in the form of Initial Noise Optimization (\textsc{InitNO}), a paradigm that refines this noise. Considering text prompts, not all random noises are effective in synthesizing semantically-faithful images. We design the cross-attention response score and the self-attention conflict score to evaluate the initial noise, bifurcating the initial latent space into valid and invalid sectors. A strategically crafted noise optimization pipeline is developed to guide the initial noise towards valid regions. Our method, validated through rigorous experimentation, shows a commendable proficiency in generating images in strict accordance with text prompts. Our code is available at \url{https://github.com/xiefan-guo/initno}.
\end{abstract}
\section{Introduction}
\label{sec:introduction}

Text-to-Image Synthesis (T2I) stands at the forefront of cutting-edge research, dedicated to the generation of authentic and visually cohesive images from text prompts. In the domain of generative models, including generative adversarial networks \cite{goodfellow2014generative,karras2019style,karras2020analyzing,karras2020training,karras2021alias}, variational autoencoders \cite{kingma2013auto}, and autoregressive models \cite{van2016conditional,vaswani2017attention,chen2020generative,esser2021taming}, diffusion models \cite{ho2020denoising,dhariwal2021diffusion} have ascended as a predominant solution. The integration of large-scale vision-language models \cite{ramesh2022hierarchical,saharia2022photorealistic,yu2022scaling,rombach2022high,kang2023scaling,balaji2022ediffi,xue2023raphael,gu2023matryoshka,feng2023dreamoving} has propelled noteworthy advancements in the text-to-image domain.

\begin{figure}
\centering
\setlength{\belowcaptionskip}{-0.2cm}
\setlength{\abovecaptionskip}{0.2cm}
\includegraphics[width=1.\linewidth]{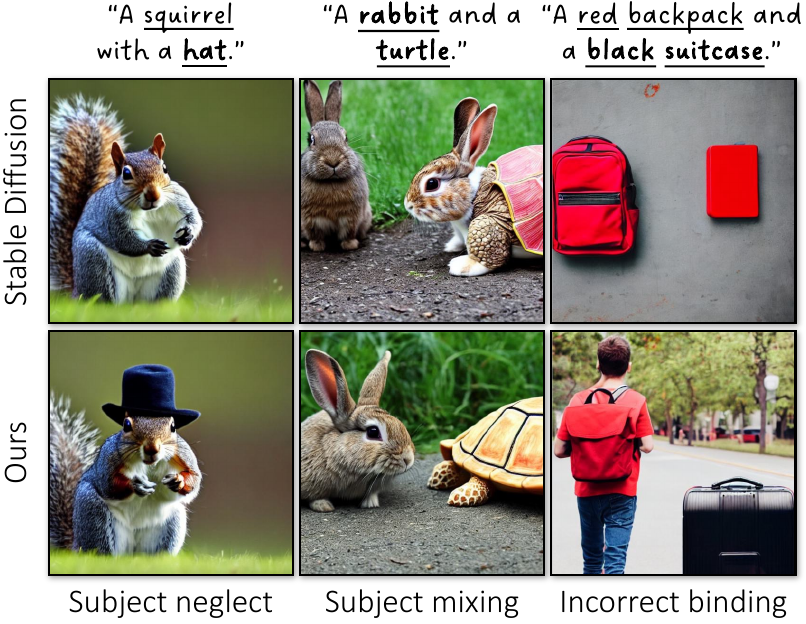}
\caption{\textbf{Example results synthesized by SD and ours.}}
\label{fig:issues}
\end{figure}

Despite the training of state-of-the-art T2I diffusion models on large-scale text-image datasets, synthesizing images precisely aligned with given text prompts remains a considerable challenge. Well-documented issues, \emph{i.e.}, subject neglect, subject mixing, and incorrect attribute binding, as illustrated in Fig.~\ref{fig:issues}, persist. We attribute these challenges to the presence of invalid initial noise.

\begin{figure*}
\centering
\setlength{\belowcaptionskip}{-0.2cm}
\setlength{\abovecaptionskip}{0.2cm}
\includegraphics[width=1.\linewidth]{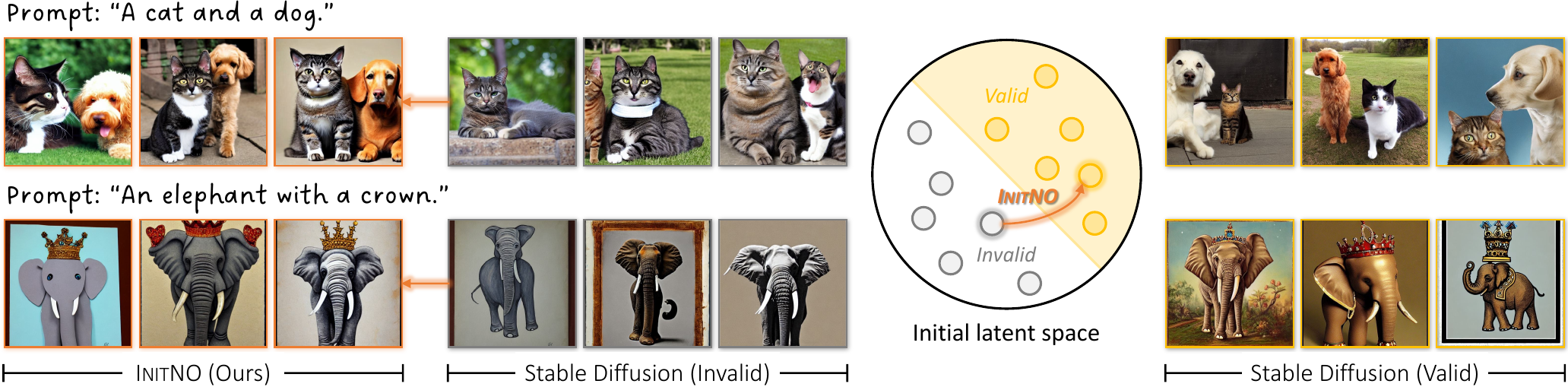}
\caption{\textbf{\textsc{InitNO}.} Our investigation dives into the exploration of various random noise configurations and their subsequent influence on the generated results. Notably, when different noises are input into SD under identical text prompts, there are marked discrepancy in the alignment between the generated image and the given text. Unsuccessful cases are delineated by gray contours, while successful instances are indicated by yellow contours. This observation underscores the pivotal role of initial noise in determining the success of the generation process. Based on this insight, we divide the initial noise space into valid and invalid regions. Introducing Initial Noise Optimization (\textsc{InitNO}), identified as orange arrow, our method is capable of guiding any initial noise into the valid region, thereby synthesizing high-fidelity results (orange contours) that precisely correspond to the given prompt. The same location employs the same random seed.}
\label{fig:motivation}
\end{figure*}

When different noise inputs are introduced to the T2I diffusion model with identical text prompts, a substantial discrepancy is observed in the alignment between images and the provided text, as depicted in Fig.~\ref{fig:motivation}. This observation signifies that not all randomly sampled noise can produce visually-consistent images. Depending on the consistency between the generated image and the target text, the initial latent space can be divided into valid and invalid regions. Noise sourced from valid regions, when input into the T2I diffusion model, results in semantically-reasonable image. Consequently, our aim is to direct any initial noise towards the valid region, thereby facilitating the generation of visually-coherent image that faithfully aligns with the input prompt.

The core of this paper lies in the concept of Initial Noise Optimization (\textsc{InitNO}), which encompasses initial latent space partitioning and the noise optimization pipeline, responsible for defining valid regions and directing initial noise navigation, respectively.

\begin{figure}
\centering
\setlength{\belowcaptionskip}{-0.4cm}
\setlength{\abovecaptionskip}{0.2cm}
\includegraphics[width=1.\linewidth]{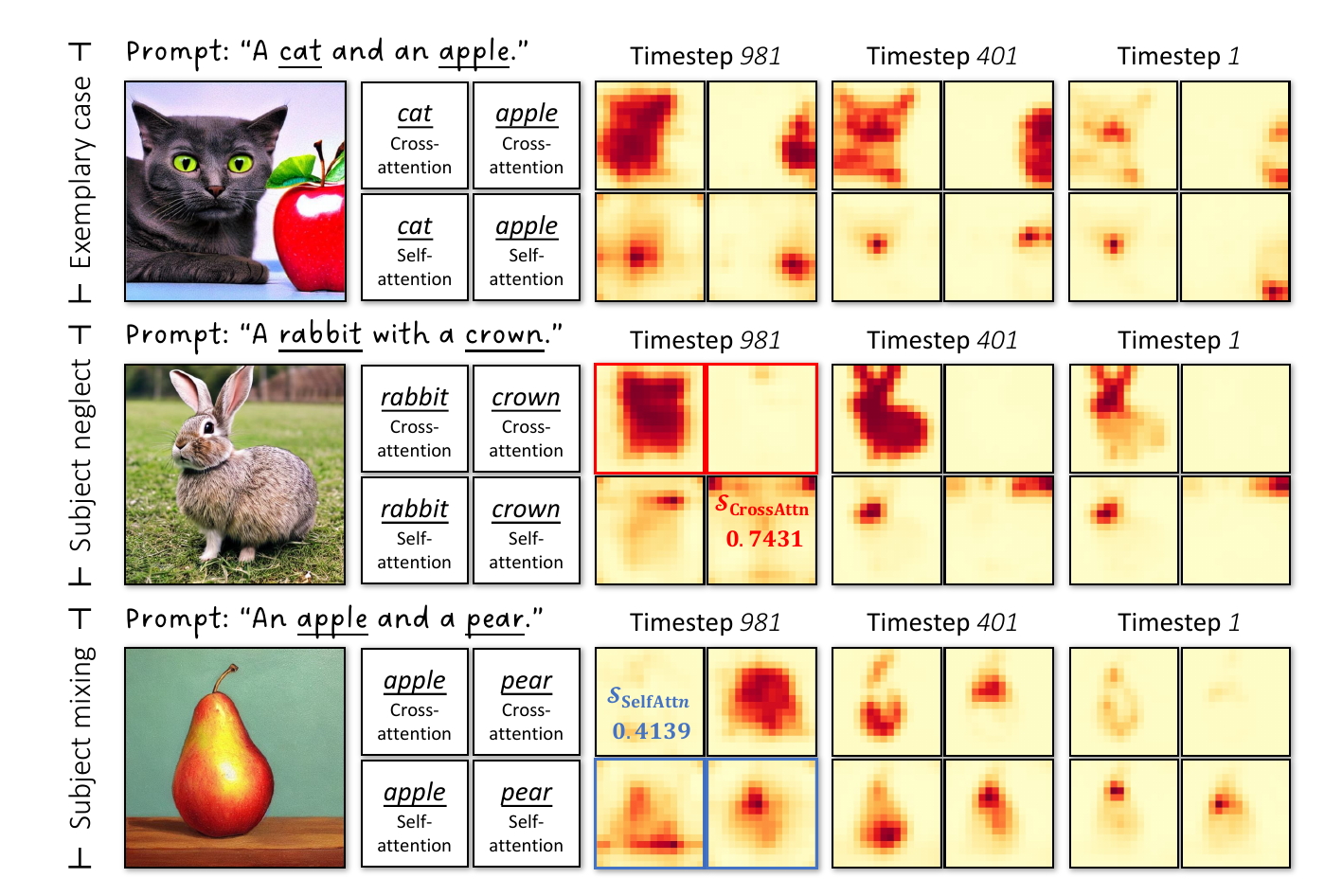}
\caption{\textbf{Visualization of the attention maps.}}
\label{fig:visualization-cross-self-attention-map}
\end{figure}

Our study thoroughly examines attention layers in the diffusion model, specifically the cross-attention map and the self-attention map. The former captures the correlation between text and image features, while the latter captures the correlation among different spatial positions within the image features. These maps effectively measure subject neglect and subject mixing, respectively (see Fig.~\ref{fig:visualization-cross-self-attention-map}). With these maps, we design the cross-attention response score and the self-attention conflict score. By setting score thresholds, we achieve further partitioning of the initial latent space. Interestingly, we find that accurate rendering of the subject fosters correct attribute binding.

The primary challenge in initial noise navigation is to balance under-optimization and over-optimization of noise. While under-optimization is still plagued by misalignment, over-optimization risks deviating from the initial distribution, \emph{i.e.}, standard Gaussian distribution, resulting in distorted images (see Fig.~\ref{fig:ablation-distribution-alignment-loss}). To address this challenge, we introduce a carefully-crafted noise optimization pipeline. With the explicit modeling of the initial latent space, a novel distribution alignment loss is incorporated to ensure the optimized noise adheres to the initial distribution.

We establish the superiority of the proposed method over state-of-the-art approaches in generating semantically accurate images. Furthermore, our method is plug-and-play and seamlessly integrates into existing diffusion models, enabling training-free controllable generation and improving alignment with specific conditions, such as layout-to-image generation (see Sec.~\ref{sec:grounded-text-to-image}).

\section{Related Work}
\label{sec:related_work}

Text-to-image synthesis strives to generate visually-realistic image that reflects given text. Early research mainly focused on GANs \cite{zhang2017stackgan,xu2018attngan,zhu2019dm,zhang2021cross,tao2022df} and autoregressive models \cite{ramesh2021zero,ding2021cogview,yu2022scaling,chang2023muse}. Recently, diffusion models \cite{ho2020denoising,dhariwal2021diffusion} have taken over the mainstream, achieving impressive results. The incorporation of large-scale vision-language models \cite{saharia2022photorealistic,yu2022scaling,rombach2022high,kang2023scaling,gu2023matryoshka} has fostered significant advancements in the text-to-image domain. Despite synthesizing highquality images, it remains challenging to produce results that properly comply with the given text prompt.

To deal this problem, some works \cite{ramesh2022hierarchical,saharia2022photorealistic,balaji2022ediffi,xue2023raphael,segalis2023picture} expand the network architecture and introduce large language models \cite{raffel2020exploring} to better hint embeddings. However, these works require training new text-to-image models and are not applicable to existing widely used models. Another line of work explores training-free improvement strategies, consistent with our approach.

Specifically, Liu \emph{et al.} \cite{liu2022compositional} propose to use multiple diffusion models for different concepts and then composite the outputs to obtain the final result. This method solves the mentioned problem to some extent, but still suffers from subject mixing .  Feng \emph{et al.} \cite{feng2023training} propose to split the input text prompt and process the cross attention layers to align the tokens and the outputs. Chefer \emph{et al.} \cite{chefer2023attend} introduce the concept of Generative Semantic Nursing (GSN), and slightly shifts the noisy image at each timestep of the denoising process, where the semantic information from the text prompt is better considered. Agarwal \emph{et al.} \cite{agarwal2023star} and Li \emph{et al.} \cite{li2023divide} further improve the optimization objective on this paradigm for better update direction. However, updating the noisy image at each denoising step requires carefully-designed optimization parameters, posing a delicate balance between under-optimization and over-optimization.

Different from previous methods, we explore a novel path by adjusting the sampled noise in the initial latent space, where full optimization is performed. It cleverly avoids the trade-off between under-optimization and over-optimization, is proven to be efficient and orthogonal to existing methods.

\section{Preliminaries}
\label{sec:preliminaries}

\noindent \textbf{Stable Diffusion Model.} This study focuses on the state-of-the-art Stable Diffusion model (SD) \cite{rombach2022high}, belonging to the category of latent diffusion models (LDMs). Unlike traditional diffusion-based approaches operating directly in the image domain, SD functions within the latent space of an autoencoder. Specifically, an encoder $\mathcal{E}\left(\cdot\right)$ learns the mapping from an input image $\mathbf{x}\in \mathcal{X}$ to a latent code $\mathbf{z} = \mathcal{E}\left( \mathbf{x} \right)$. Subsequently, a decoder $\mathcal{D}(\cdot)$ reconstructs the input image, aiming for $\mathcal{D}\left( \mathcal{E}\left( \mathbf{x} \right) \right) \approx \mathbf{x}$.

Once the autoencoder is trained, a Denoising Diffusion Probabilistic Model (DDPM) \cite{ho2020denoising} operates within the acquired latent space to generate a denoised version of an input latent $\mathbf{z}_t$ at each timestep $t$. During denoising, the diffusion model can be conditioned on additional inputs. In the case of SD, this condition typically takes the form of a text embedding generated by a pre-trained CLIP text encoder \cite{radford2021learning}. The conditioning embedding for the given prompt $\mathbf{y}$ is denoted as $\mathbf{c} = f_{\text{CLIP}}\left( \mathbf{y} \right)$. The DDPM model $\epsilon_\theta(\cdot)$, parametrized by $\theta$, optimizes the following loss:
\begin{equation}
\label{eq:ddpm}
    \mathcal{L} = \mathbb{E}_{\mathbf{z}\sim\mathcal{E}\left( \mathbf{x} \right), \mathbf{c}, \epsilon \sim \mathcal{N}\left( \mathbf{0}, \mathbf{1} \right), t}\left[ \| \epsilon - \epsilon_{\theta}\left( \mathbf{z}_t, \mathbf{c}, t \right) \|_2^2 \right],
\end{equation}

During inference, a latent variable $\mathbf{z}_T$ is sampled from the standard Gaussian distribution $\mathcal{N}(\mathbf{0}, \mathbf{1})$ and subjected to sequential denoising procedures of DDPM to derive a refined latent representation $\mathbf{z}_0$. This denoised latent $\mathbf{z}_0$ is then fed into the decoder $\mathcal{D}(\cdot)$ to synthesize the corresponding image $\mathcal{D}\left(\mathbf{z}_0\right)$.

\noindent \textbf{Cross-attention layer.} Text-image correspondence in SD is achieved through the cross-attention layer, enabling text condition guidance. Specifically, the pre-trained CLIP text encoder \cite{radford2021learning} embeds the text prompt $\mathbf{y} = \{\mathbf{y}_1, \mathbf{y}_2, \cdots, \mathbf{y}_n \}$ into a sequential embedding, serving as the condition $\mathbf{c} = f_{\text{CLIP}}\left( \mathbf{y} \right)$. Linear projections are then employed to extract the key $\mathbf{K}$ and value $\mathbf{V}$ from $\mathbf{c}$,  while the query $\mathbf{Q}$ is mapped from the intermediate features of UNet. The cross-attention map $\mathbf{A}^c$ is computed via:
\begin{equation}
\label{eq:cross-attention}
   \mathbf{A}^c = \text{softmax}\left( \frac{\mathbf{Q}\mathbf{K}^T}{\sqrt{d}} \right),
\end{equation}
where $d$ is the channel dimension. For ease of representation, we omit the denoising timestep $t$. We denote the attention map that corresponds to the $i$-th text token as $\mathbf{A}^c_{\mathbf{y}_i}$, and $\mathbf{A}^c_{\mathbf{y}_{i}}\left[x, y\right]$ is the probability assigned to text token $\mathbf{y}_i$ for the $(x,y)$-th spatial patch of the intermediate feature map.

\noindent \textbf{Self-attention layer.} In the self-attention layer, intermediate features serve as the key $\mathbf{K}$ and value $\mathbf{V}$ instead of the text condition $\mathbf{c}$, allowing synthesis of globally coherent structures by correlating image tokens across diverse regions. The self-attention map $\mathbf{A}^s$ is computed using the same formula as Eq.~\ref{eq:cross-attention}, and $\mathbf{A}^s_{x,y}$ denotes the attention map that corresponds to the $(x,y)$-th spatial patch.

\section{\textsc{InitNO}}
\label{sec:approach}

\begin{figure*}
\centering
\setlength{\belowcaptionskip}{-0.2cm}
\setlength{\abovecaptionskip}{0.2cm}
\includegraphics[width=1.\linewidth]{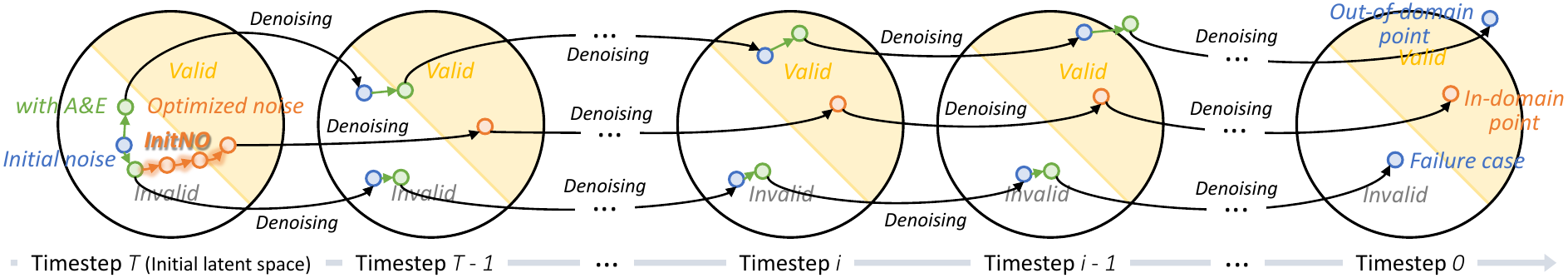}
\caption{\textbf{Comparison of Attend-and-Excite with our method.} Attend-and-Excite suffers from the trade-off between under-optimization and over-optimization. Under-optimization (\emph{lower path}) remains confined to invalid regions, while over-optimization (\emph{upper path}) carries the risk of deviating from the distribution of diffusion model. Our approach (\emph{middle path}) skillfully addresses this challenge by prioritizing noise optimization in the initial latent space, ensuring sufficient and appropriate optimization.}
\label{fig:initno-method}
\end{figure*}

\begin{figure}
\centering
\setlength{\belowcaptionskip}{-0.2cm}
\setlength{\abovecaptionskip}{0.2cm}
\includegraphics[width=1.\linewidth]{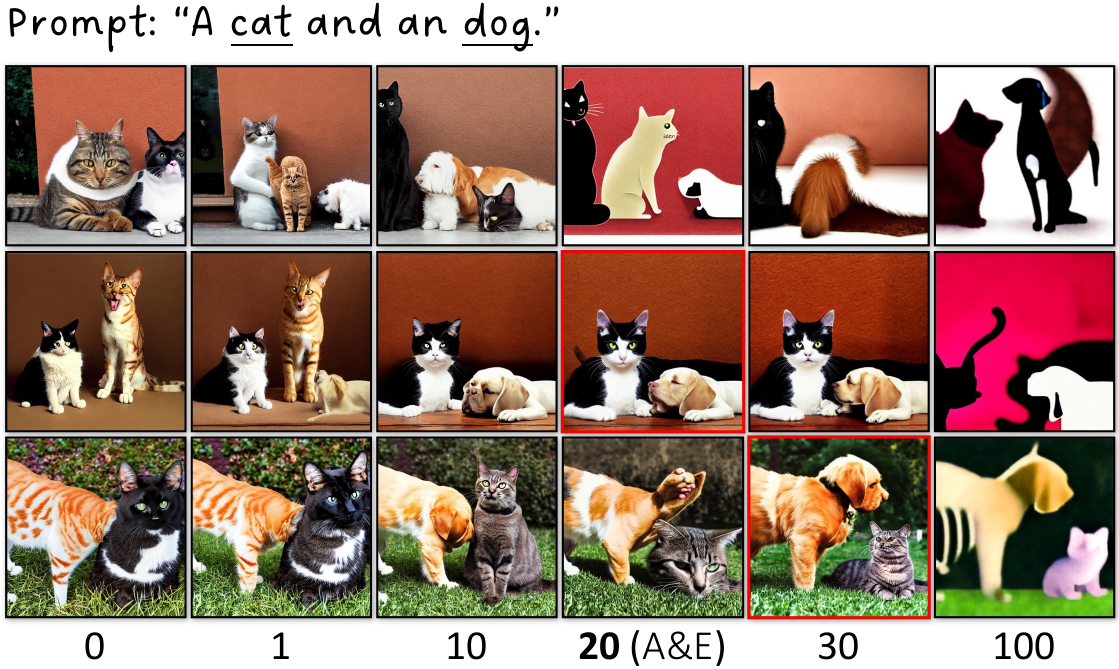}
\caption{\textbf{Effect of the scale factor of Attend-and-Excite.} Given the same text prompt, we adjust the scale factor of Attend-and-Excite \cite{chefer2023attend}, 20 is used in the original work. Images of varying quality are synthesized, and the red box indicates the highest quality image. Images on the same row share an identical random seed.}
\label{fig:attend-and-excite-scale-factor}
\end{figure}

The core of our methed is the Initial Noise Optimization (\textsc{InitNO}), which comprises the initial latent space partitioning and the noise optimization pipeline, responsible for defining valid regions and steering noise navigation, respectively. This section provides detailed explanations of both components in Sec.~\ref{sec:initial_latent_space_partitioning} and~\ref{sec:noise-optimization-pipeline}.

\subsection{Initial latent space partitioning}
\label{sec:initial_latent_space_partitioning}

Prevailing studies \cite{hertz2023prompt,tumanyan2023plug,chefer2023attend,parmar2023zero,cao2023masactrl} demonstrate the informativeness of cross-attention maps, elucidating their capability to partially quantify the alignment between generated image and provided text. In this work, we extend these findings by thoroughly investigating the self-attention maps, which capture inter-spatial correlations within image features. We argue that cross-attention maps and self-attention maps can inherently gauge subject neglect and subject mixing, respectively. Leveraging these maps, we formulate the cross-attention response score and the self-attention conflict score. By establishing the score threshold, we achieve the partitioning of the initial latent space.

\noindent \textbf{Cross-attention response score.} We follow \cite{chefer2023attend} and derive the final cross-attention map by averaging the cross-attention maps in all layers and heads with a resolution of 16$\times$16 pixels. The aggregated map $\mathbf{A}^c\in \mathbb{R}^{16\times 16\times n}$ contains $n$ spatial attention maps, each corresponding to a token of the text prompt. We also re-weight the attention values by ignoring the attention of the specialized token $\mathsf{sot}$ and performing a softmax operation on the remaining tokens. Fig.~\ref{fig:visualization-cross-self-attention-map} shows the cross-attention map for each subject token, reflecting the spatial layout information corresponding to the generated image. Notably, a proficient denoising process can assign sufficiently high cross-attention responses to each subject during the early stages.

Conversely, a low attention response can lead to the absence of the target subject, \emph{i.e.}, {subject neglect}. As depicted in the red box in Fig.~\ref{fig:visualization-cross-self-attention-map}, the token \emph{\underline{crown}} registers a significantly lower response than \emph{\underline{rabbit}} in the initial denoising step, and \emph{\underline{crown}} is missing from the generated image. Building upon the insights from \cite{chefer2023attend}, we define the cross-attention response score as:
\begin{equation}
\label{eq:cross-attention-score}
    \mathcal{S}_\text{CrossAttn} = 1 - \min_{\mathbf{y}_i\in \mathcal{Y}} \max \left(\mathbf{A}^c_{\mathbf{y}_i}\right),
\end{equation}
where $\mathcal{Y}$ denotes the set of target tokens. Specifically, in this work, $\mathcal{S}_\text{CrossAttn}$ is calculated over the initial noise and the first denoising step. We further define the cross-attention response score threshold $\tau_c$, empirically set to 0.2 in our experiments, where scores below this threshold are classified as valid, and those above are deemed invalid.

\noindent \textbf{Self-attention conflict score.} Consistent with the cross-attention map extraction strategy, we extract the final self-attention map $\mathbf{A}^s\in \mathbb{R}^{16\times16\times(16\times16)}$. For the subject token $\mathbf{y}_i$ of interest, we first query the spatial coordinates $(x_i, y_i)$ corresponding to its maximum cross-attention value:
\begin{equation}
\label{eq:self-attention-coordinate}
   x_i, y_i = \mathop{\arg\max}\limits_{x, y}\ \mathbf{A}_{\mathbf{y}_i}^c \left[ x, y \right].
\end{equation}

Subsequently, we obtain its corresponding self-attention map $\mathbf{A}_{x_i, y_i}^s$, as shown in Fig.~\ref{fig:visualization-cross-self-attention-map}. For an exemplary denoising process, the self-attention maps of different subject tokens can be distinctly partitioned in the spatial domain. However, existing diffusion models, \emph{e.g.}, SD, suffer from self-attention map overlap, leading to a failure case of subject mixing. As indicated in the blue box in Fig.~\ref{fig:visualization-cross-self-attention-map}, the self-attention maps of \emph{\underline{apple}} and \emph{\underline{pear}} significantly overlap, resulting in a combination of both. To quantify this phenomenon, we introduce the self-attention conflict score:
\begin{equation}
\begin{aligned}
\label{eq:self-attention-score}
    \mathcal{S}_\text{SelfAttn} &= \sum_{\mathbf{y}_i, \mathbf{y}_j \in \mathcal{Y}, \forall i < j} \frac{\text{f}\left( \mathbf{y}_i, \mathbf{y}_j \right)}{N}, \\
    \text{f}\left( \mathbf{y}_i, \mathbf{y}_j \right) &= \frac{\sum_{x,y} \min \left( \mathbf{A}_{x_i, y_i}^s\left[x,y\right], \mathbf{A}_{x_j, y_j}^s\left[x,y\right] \right)}{\sum_{x,y} \left( \mathbf{A}_{x_i, y_i}^s\left[x,y\right] + \mathbf{A}_{x_j, y_j}^s\left[x,y\right] \right)},
\end{aligned}
\end{equation}
where $N$ is the number of pairs of $\mathbf{y}_i$ and $\mathbf{y}_j$. $\mathcal{S}_\text{SelfAttn}$ is also calculated based on the initial noise and the first denoising step. We further define the self-attention conflict score threshold $\tau_s$, set to 0.3 in our experiments. Scores below $\tau_s$ are classified as valid, and those above as invalid.

Finally, we define the initial noise that passes both the cross-attention response score test and the self-attention conflict test as valid noise, and vice versa as invalid.

\subsection{Noise optimization pipeline}
\label{sec:noise-optimization-pipeline}

As illustrated in Fig.~\ref{fig:initno-method}, existing methods most relevant to ours, represented by Attend-and-Excite \cite{chefer2023attend}, modify the noisy image at each denoising step, and encourage it to better encapsulate the semantic information from text. However, they necessitate meticulously calibrated optimization parameters. As shown in Fig.~\ref{fig:attend-and-excite-scale-factor}, different optimization parameter settings directly affect the final generated image. Under-optimization leads to subpar quality, while over-optimization skews the noisy image away from the distribution of SD, resulting in an out-of-domain image. Moreover, even with the same text prompt, optimal parameters vary with noise, making case-specific tuning a laborious task.

We bypass this challenge by prioritizing noise optimization in the initial latent space, ensuring sufficient and appropriate adjustment. Specifically, \textsc{InitNO} is exclusively dedicated to navigating within the initial latent space, moving randomly sampled initial noise into the valid region. Unlike denoising processes that implicitly model the noisy image distribution, the initial latent space typically adheres to a standard Gaussian distribution, providing an avenue to restrict the optimized noise to the domain.

\noindent \textbf{Noise update strategy.} Given a noise $\epsilon \in \mathcal{N}\left( \mathbf{0}, \mathbf{1} \right)$, as opposed to the commonly used incremental update: ${\epsilon}{'} \gets \epsilon + \Delta \epsilon$, we adopt a distribution optimization strategy to update the noise: 
\begin{align}
\label{eq:distribution-optimization}
    {\mu}{'} \gets \mu + \Delta\mu, \qquad
    \sigma' \gets \sigma + \Delta\sigma,
\end{align}
where $\mu$ and $\sigma$ are the mean and standard deviation of the Gaussian distribution, initialized to $\mathbf{0}$ and $\mathbf{1}$, respectively. Thus, we have: $\epsilon' \gets \mu' + \sigma'\epsilon$, and $\epsilon' \sim \mathcal{N}\left( \mu', {\sigma'}^2 \right)$.

\noindent \textbf{Loss functions.} We employ a joint loss for optimizing initial noise, comprising cross-attention response loss, self-attention conflict loss, and distribution alignment loss. Intuitively, the first two are formulated as:
\begin{align}
\label{eq:cross-self-loss}
    \mathcal{L}_{\text{CrossAttn}} = \mathcal{S}_{\text{CrossAttn}},\qquad \mathcal{L}_{\text{SelfAttn}} = \mathcal{S}_{\text{SelfAttn}}.
\end{align}

The distribution alignment loss utilizes Kullback-Leibler divergence \cite{kingma2013auto} to constrain the consistency between the target Gaussian distribution $\mathcal{N}\left(\mu, \sigma^2\right)$ and the standard Gaussian distribution $\mathcal{N}\left(\mathbf{0}, \mathbf{1}\right)$:
\begin{equation}
\begin{aligned}
\label{eq:distribution-alignment-loss}
    \mathcal{L}_{\text{KL}} &= \textsf{KL}\left( \mathcal{N}\left(\mu, \sigma^2\right)\|\  \mathcal{N}\left(\mathbf{0}, \mathbf{1}\right) \right).
\end{aligned}
\end{equation}

In summary, the joint loss is written as:
\begin{equation}
\begin{aligned}
\label{eq:joint-loss}
    \mathcal{L}_{\text{joint}} &= \lambda_{{1}}\mathcal{L}_{\text{CrossAttn}} + \lambda_{{2}}\mathcal{L}_{\text{SelfAttn}} + \lambda_{{3}}\mathcal{L}_{\text{KL}},
\end{aligned}
\end{equation}
where we empirically set $\lambda{{1}}$ = $1$, $\lambda_{{2}}$ = $1$, and $\lambda_{{3}}$ = $500$.

\noindent \textbf{Noise optimization procedure.} The outlined noise optimization procedure is encapsulated in Algorithm~\ref{algorithm:noise-optimization-procedure}. Sampling the initial noise $\mathbf{z}_T \sim \mathcal{N}\left(\mathbf{0}, \mathbf{1}\right)$, we initialize the learnable parameters $\mu$ to $\mathbf{0}$ and $\sigma$ to $\mathbf{1}$. We optimize these parameters using the joint loss, ensuring that the optimized noise $\mu + \sigma\mathbf{z}^T$ resides within the valid initial latent space. The optimization employs the Adam optimizer with a learning rate of $1\times10^{-2}$.

\begin{algorithm2e}
\DontPrintSemicolon
\KwIn {A pre-trained T2I diffusion model $\text{SD}\left(\cdot \right)$, a text prompt $\mathbf{y}$, a noise $\mathbf{z}_T$, threshold: $\tau_c$, $\tau_s$, $\tau_{\text{MaxStep}}$, and $\tau_{\text{MaxRound}}$.}
\KwOut {Generated image $\mathbf{x}$.}
Noise pool $\mathcal{P}$ $\gets$ $\{\}$\;
\For{\emph{$i$ $=$ $1$ {to} $\tau_{\text{MaxRound}}$}}{
    {Initialize} $\mathbf{z}_T\sim \mathcal{N}\left(\mathbf{0}, \mathbf{1}\right)$, $\mu$ $\gets$ $\mathbf{0}$, $\sigma$ $\gets$ $\mathbf{1}$\;
    \For{\emph{$j$ $=$ $1$ {to} $\tau_{\text{MaxStep}}$}}{
        $\_$, $\mathbf{A}^c$, $\mathbf{A}^s$ $\gets$ $\text{SD}\left(\mu + \sigma \mathbf{z}^T, \mathbf{y} \right)$\;
        Calculate $\mathcal{S}_{\text{CrossAttn}}$ (Eq.~\ref{eq:cross-attention-score}), $\mathcal{S}_{\text{SelfAttn}}$ (Eq.~\ref{eq:self-attention-score})\;
        \eIf{\emph {$\mathcal{S}_{\text{CrossAttn}}$ $<$ $\tau_c$ {and} $\mathcal{S}_{\text{SelfAttn}}$ $<$ $\tau_s$}}{
            $\hat{\mathbf{z}}^T$ $\gets$ $\mu$ $+$ $\sigma \mathbf{z}^T$\;
            {\small $\mathsf{Go\ to\ step\ 15}$}\Comment{$\mathsf{valid\ noise}$}\; 
        } {
            Calculate $\mathcal{L}_{\text{joint}}$ (Eq.~\ref{eq:joint-loss})\;
            $\mu$, $\sigma$ $\gets$ {$\text{Adam}$}$\left( \mu, \sigma, \mathcal{L}_{\text{joint}} \right)$\;
        }
    }
    Add $\mu$ $+$ $\sigma \mathbf{z}^T$ to $\mathcal{P}$\;
}
$\hat{\mathbf{z}}^T$ $\gets$ $\mathop{\arg\min}\limits_{\mathbf{z}^T\in \mathcal{P}}\ \mathcal{S}_{\text{CrossAttn}}$ $+$ $\mathcal{S}_{\text{SelfAttn}}$\;
$\mathbf{x}$, $\_$, $\_$ $\gets$ $\text{SD}\left( \hat{\mathbf{z}}^T, \mathbf{y} \right)$ {or} $\text{AE}\left( \hat{\mathbf{z}}^T, \mathbf{y} \right)$\;
\Return $\mathbf{x}$\;
\caption{\bf \textsc{InitNO}}
\label{algorithm:noise-optimization-procedure}
\end{algorithm2e}

To mitigate computational inefficiency due to excessively challenging samples, we impose a maximum limit on optimization iterations, denoted as $\tau_{\text{MaxStep}}$. If we fail to reach the valid region after $\tau_{\text{MaxStep}}$ updates, we resample $\mathbf{z}_T$, reset $\mu$ and $\sigma$, and commence a new round of the optimization process. We also define a maximum limit on optimization rounds, denoted as $\tau_{\text{MaxRound}}$. In our experiments, we set $\tau_{\text{MaxStep}}$ and $\tau_{\text{MaxRound}}$ as 50 and 5, respectively. For a minority of intricate texts where effective noise cannot be found within affordable optimization rounds, we opt for the noise that minimizes the sum of $\mathcal{S}_{\text{CrossAttn}}$ and $\mathcal{S}_{\text{SelfAttn}}$ from the existing noise pool $\mathcal{P}$.

\section{Experiments}
\label{sec:experiments}

\begin{figure*}
\centering
\setlength{\belowcaptionskip}{-0.2cm}
\setlength{\abovecaptionskip}{0.2cm}
\includegraphics[width=1.\linewidth]{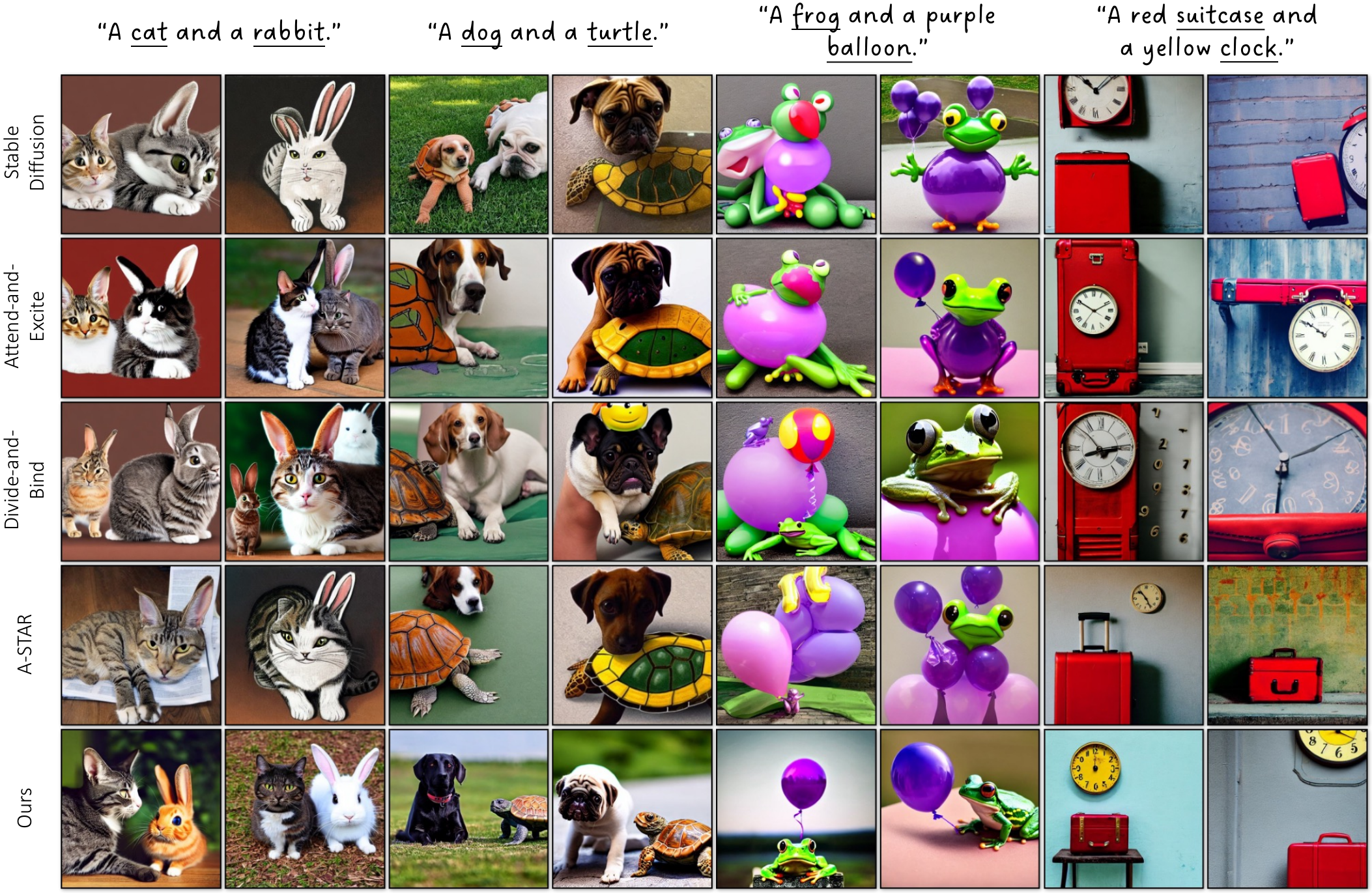}
\caption{\textbf{Qualitative comparison.} Each image is generated with the same text prompt and random seed for all methods. The subject tokens are highlighted in \underline{underline}. Our method shows excellent alignment with text prompts while maintaining a high level of realism.}
\label{fig:qualitative-comparison}
\end{figure*}

\subsection{Experimental settings} 

\noindent \textbf{Implementation Details.} Following \cite{chefer2023attend}, we utilize the official Stable Diffusion v1.4 text-to-image model. A fixed guidance level of 7.5 is employed. We apply a Gaussian filter with a kernel size of 3 and a standard deviation of 0.5 to smooth the cross-attention map $\mathbf{A}^c$ and the self-attention map $\mathbf{A}^s$. For the denoising process, $T$ is set to 50. The target token identification process can be performed either manually or automatically with the assistance of GPT \cite{brown2020language}.

\noindent \textbf{Datasets.} We assess our method using the Animal-Animal, Animal-Object, and Object-Object datasets \cite{chefer2023attend}, which comprise two subjects and selectively assign a color to each subject. It is noteworthy that our method is not constrained to this scenario, and it can be applied to complex prompts containing any number or type of subjects and attributes, as discussed in Sec.~\ref{sec:more-results} and supplementary material.

\subsection{Qualitative comparison}

Fig.~\ref{fig:qualitative-comparison} presents a comparative analysis of our results against state-of-the-art counterparts, under the same text prompts and random seeds. It can be seen that existing methods commonly suffer from serious problems of subject mixing. For instance, in the case of \emph{a cat and a rabbit}, conventional methods frequently yield erroneous outputs characterized by the fusion of cat faces and rabbit ears. Leveraging the self-attention conflict mechanism, our method can effectively decouple the two concepts of \emph{\underline{cat}} and \emph{\underline{rabbit}} and render them realistically and separately. Concurrently, Stable Diffusion \cite{rombach2022high}, Composable Diffusion \cite{liu2022compositional}, and Structure Diffusion \cite{feng2023training} suffer from the issue of subject neglect. Attend-and-Excite \cite{chefer2023attend}, Divide-and-Bind \cite{li2023divide}, and A-STAR \cite{agarwal2023star} adhere to the pattern of adjusting noisy images at each denoising step. Given the same text prompt, using the same adjustment strategy, the quality of results generated by sampling different noises vary significantly. For instance, for the prompt \emph{a dog and a turtle}, only the first example produced by Divide-and-Bind \cite{li2023divide} is successful. In contrast, our method exhibits better generalization and adapts to diverse prompts. Furthermore, our approach also demonstrates more faithful binding of properties, \emph{e.g.}, \emph{yellow clock}.

\begin{figure*}
\centering
\setlength{\belowcaptionskip}{-0.2cm}
\setlength{\abovecaptionskip}{0.2cm}
\includegraphics[width=1.\linewidth]{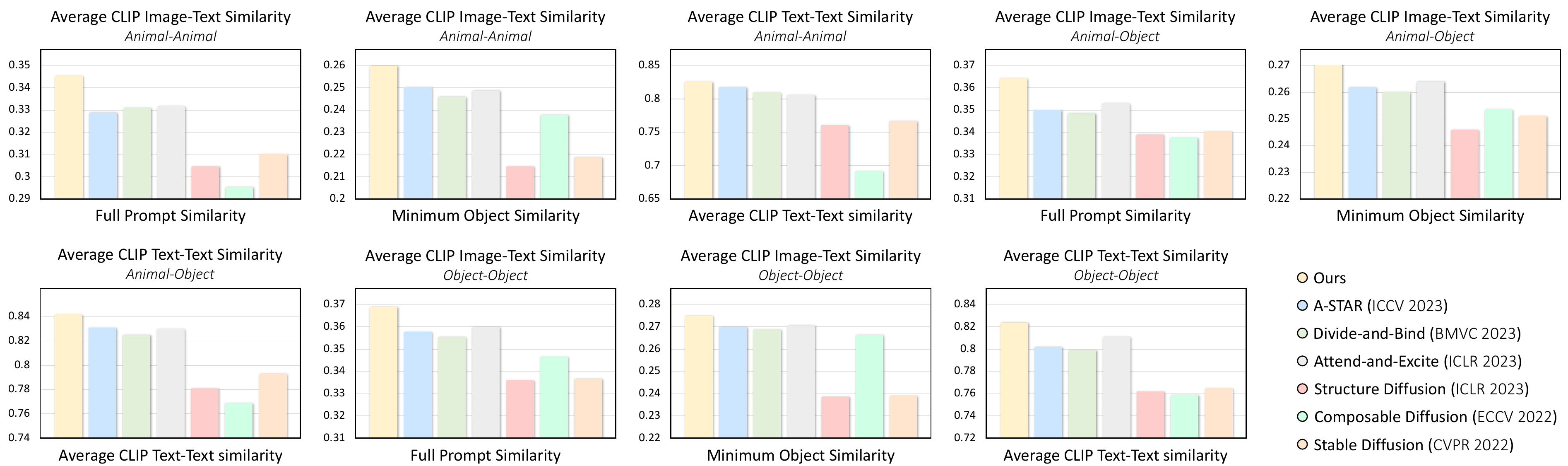}
\caption{\textbf{Objective evaluation.} Average CLIP Image-Text Similarity, including Full Prompt Similarity and Minimum Object Similarity, and Average CLIP Text-Text Similarity are reported for the quantitative measurement. \emph{Higher} is better.}
\label{fig:objective-evaluation}
\end{figure*}

\subsection{Quantitative comparison}

\noindent \textbf{Objective Evaluation.} Following to the protocol in \cite{chefer2023attend}, we quantitatively evaluate the proposed method using CLIP Image-Text Similarity and CLIP Text-Text Similarity. For each prompt, 64 images are generated, and the average CLIP cosine similarity is computed, maintaining consistent random seeds across all methods. Specifically, for CLIP Image-Text Similarity, we report full prompt similarity (cosine similarity between the full prompt and generated image) and minimum object similarity (minimum of the similarities between the generated image and each of the two subject prompts). For CLIP Text-Text Similarity, text-text similarities are computed by captioning the generated images with BLIP \cite{li2022blip} and comparing them with the input prompt. As shown in Fig.~\ref{fig:objective-evaluation}, the proposed method outperforms other approaches.

\begin{figure}
\centering
\setlength{\belowcaptionskip}{-0.2cm}
\setlength{\abovecaptionskip}{0.2cm}
\includegraphics[width=1.\linewidth]{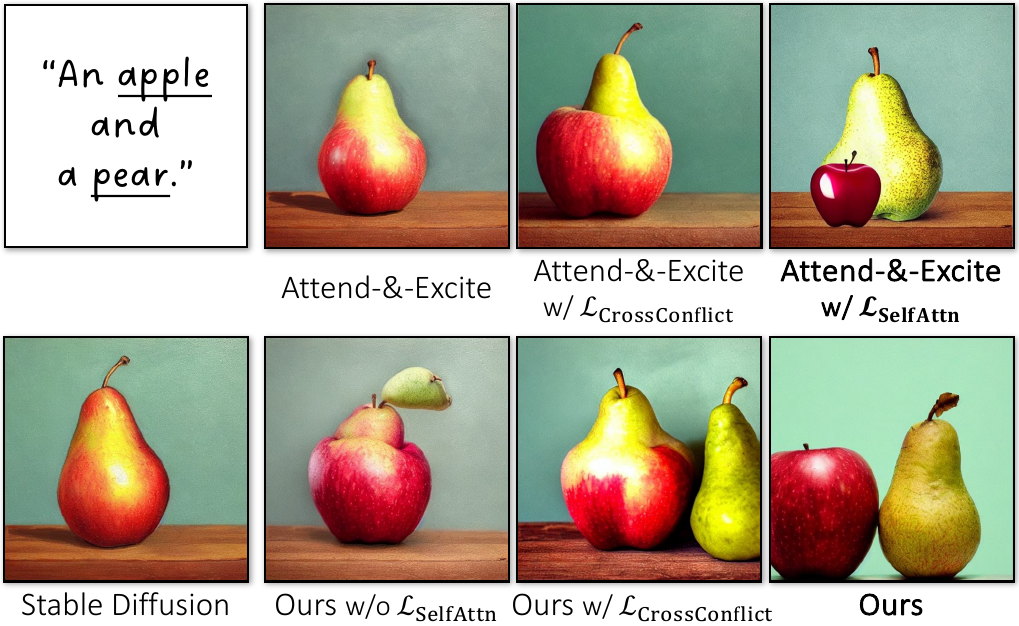}
\caption{\textbf{Visualization of effects of $\mathcal{L}_{\text{SelfAttn}}$.}}
\label{fig:ablation-self-attention-conflict-loss}
\end{figure}

\noindent \textbf{User study.} We also conduct a subjective user study involving 12 volunteers with expertise in image processing. Participants were asked to select the most visually-appealing and semantically-faithful image from those generated by our proposed method and the state-of-the-art approaches. Specifically, each participant has 20 questions. We tally the votes and show the statistics in Table~\ref{tab:user-study}. Our method performs favorably against the other methods.

\noindent \textbf{Inference time.} Evaluated on a single Tesla V100 (32GB), 100 images with a resolution of 512$\times$512 pixels are randomly generated, SD synthesizes an image in an average of 8.34 seconds, while our method takes 18.93 seconds.  

\begin{table}
\small
\centering
\setlength{\belowcaptionskip}{-0.5cm}
\setlength{\abovecaptionskip}{0.2cm}
\centering
\begin{tabular}{l|c}
\toprule
\multicolumn{1}{c|}{Method} & User study \\
\midrule
\midrule
Stable Diffusion \cite{rombach2022high} & 4.17\% \\
Composable Diffusion \cite{liu2022compositional} & 2.50\% \\
Structure Diffusion \cite{feng2023training} &  3.33\% \\
Attend-and-Excite \cite{chefer2023attend} & 14.17\% \\
Divide-and-Bind \cite{li2023divide} & 6.67\% \\
A-STAR \cite{agarwal2023star} & 5.83\% \\
\midrule
\textsc{InitNO} (Ours) & {\bf 63.33\%} \\
\bottomrule
\end{tabular}
\caption{\textbf{User study.} \textsc{InitNO} performs over other counterparts.}
\label{tab:user-study}
\end{table}

\subsection{Ablation study} 

\noindent \textbf{On self-attention conflict loss.} The self-attention conflict loss is designed to address the subject mixing issue arising from self-attention map overlap, and it can be effortlessly incorporated into existing methods. As depicted in Fig.~\ref{fig:ablation-self-attention-conflict-loss}, the self-attention conflict loss effectively addresses the mixing issue between \emph{\underline{apple}} and \emph{\underline{pear}}. We also experiment with a cross-attention conflict loss \cite{agarwal2023star}, which replaces the self-attention map with cross-attention ones, but it did not produce satisfactory results. We surmise this is due to the inaccurate text embedding extracted by CLIP. Conversely, the self-attention map offers a more intuitive solution.

\begin{figure}
\centering
\setlength{\belowcaptionskip}{-0.2cm}
\setlength{\abovecaptionskip}{0.2cm}
\includegraphics[width=1.\linewidth]{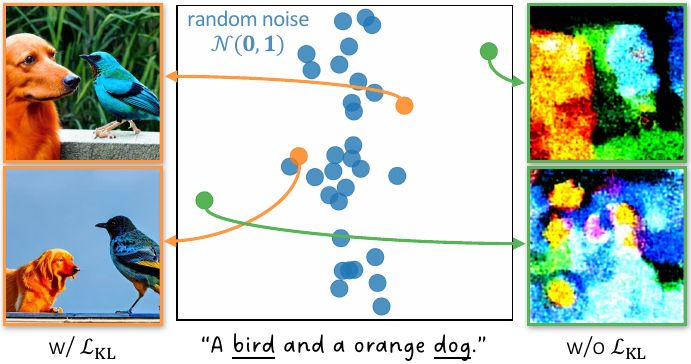}
\caption{\textbf{Visualization of effects of $\mathcal{L}_{\text{KL}}$.}}
\label{fig:ablation-distribution-alignment-loss}
\end{figure}

\begin{figure*}
\centering
\setlength{\belowcaptionskip}{-0.3cm}
\setlength{\abovecaptionskip}{0.2cm}
\includegraphics[width=1.\linewidth]{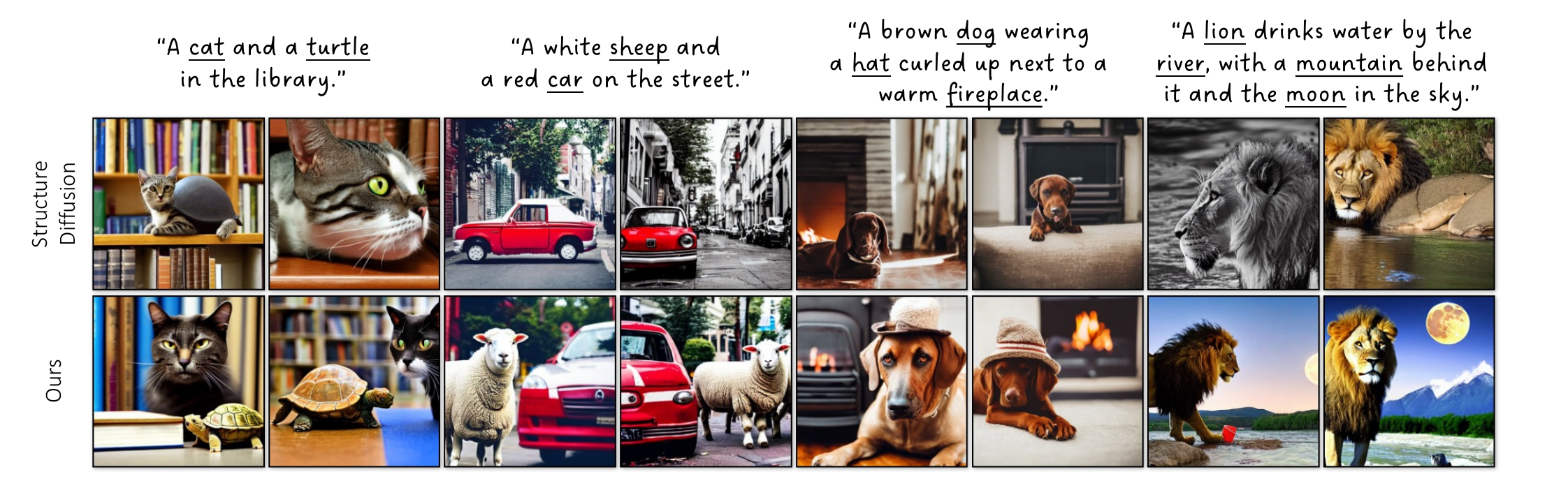}
\caption{\textbf{Qualitative comparison with complex text prompts.} Each image is generated with the same text prompt and random seed for all methods. The subject tokens are highlighted in \underline{underline}.}
\label{fig:complex-qualitative-comparison}
\end{figure*}

\begin{figure}
\centering
\setlength{\belowcaptionskip}{-0.1cm}
\setlength{\abovecaptionskip}{0.2cm}
\includegraphics[width=1.\linewidth]{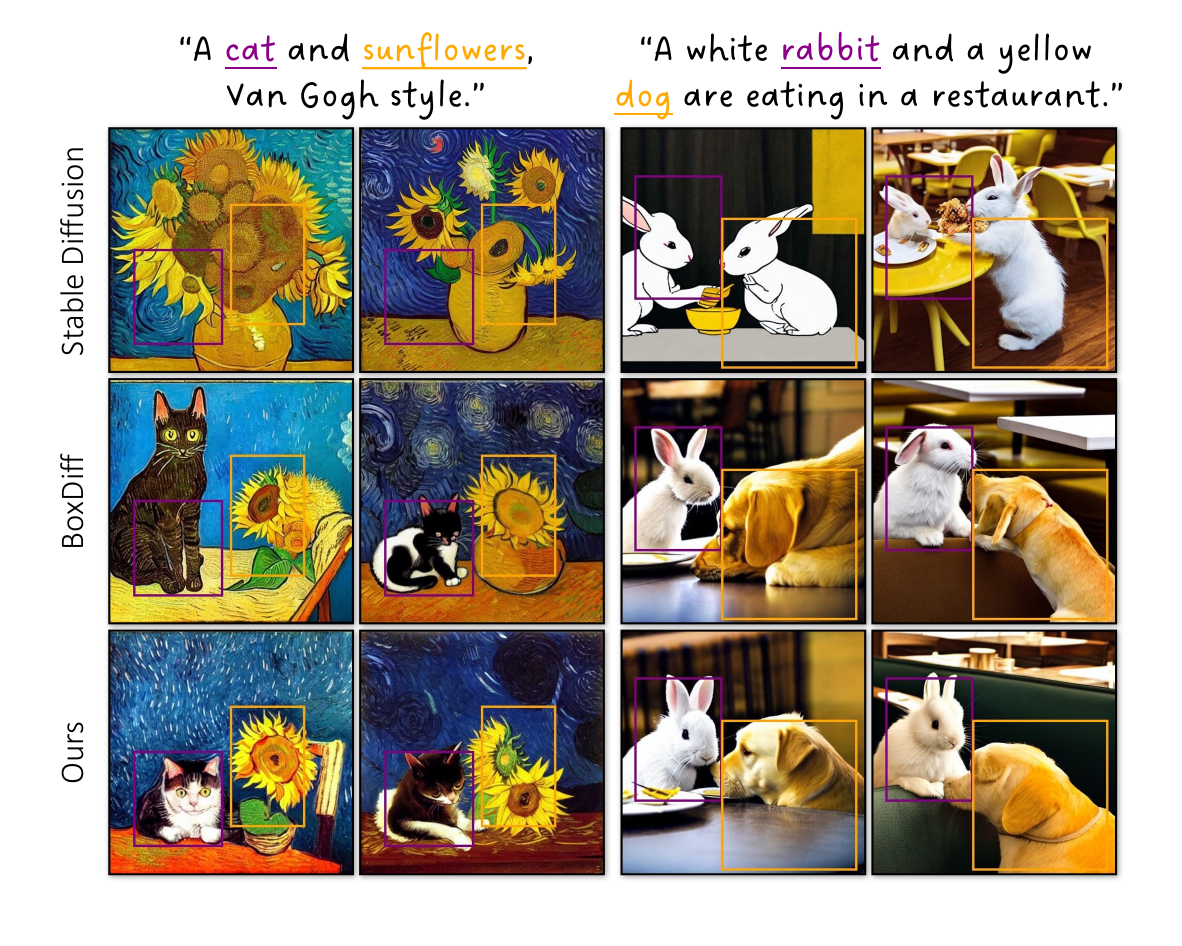}
\caption{\textbf{Grounded Text-to-Image.} Due to sufficient adjustments, \textsc{InitNO} generates more accurately location-aware \emph{\underline{cat}}.}
\label{fig:boxdiff}
\end{figure}

\noindent \textbf{On distribution alignment loss.} The distribution alignment loss is introduced to ensure that the optimized noise $\hat{\mathbf{z}}_T$ adheres to the standard Gaussian distribution. As visualized in the t-SNE plot of initial noise in Fig.~\ref{fig:ablation-distribution-alignment-loss}, without constraints, the optimized noise may deviate from the standard Gaussian distribution, resulting in distorted results.

\subsection{Grounded Text-to-Image}
\label{sec:grounded-text-to-image}

\textsc{InitNO} is a plug-and-play approach that can be effortlessly integrated with existing diffusion models to enable training-free controllable generation, such as layout-to-image, mask-to-image generation, etc. As demonstrated in Fig.~\ref{fig:boxdiff}, we incorporate \textsc{InitNO} into BoxDiff \cite{xie2023boxdiff}, where modifications to the intermediate noisy image are transferred to the initial latent space, resulting in the synthesis of faithful and location-aware images.

\subsection{More results}
\label{sec:more-results}

\noindent \textbf{Complex prompts.}  Fig.~\ref{fig:complex-qualitative-comparison} presents examples of complex prompts, inclusive of prompts with three or more subjects and intricate attributes. The synthesis of high-fidelity images is demonstrated, underscoring the capacity of our method to manage complex semantic scenes.

\begin{figure}
\centering
\setlength{\belowcaptionskip}{-0.1cm}
\setlength{\abovecaptionskip}{0.2cm}
\includegraphics[width=1.\linewidth]{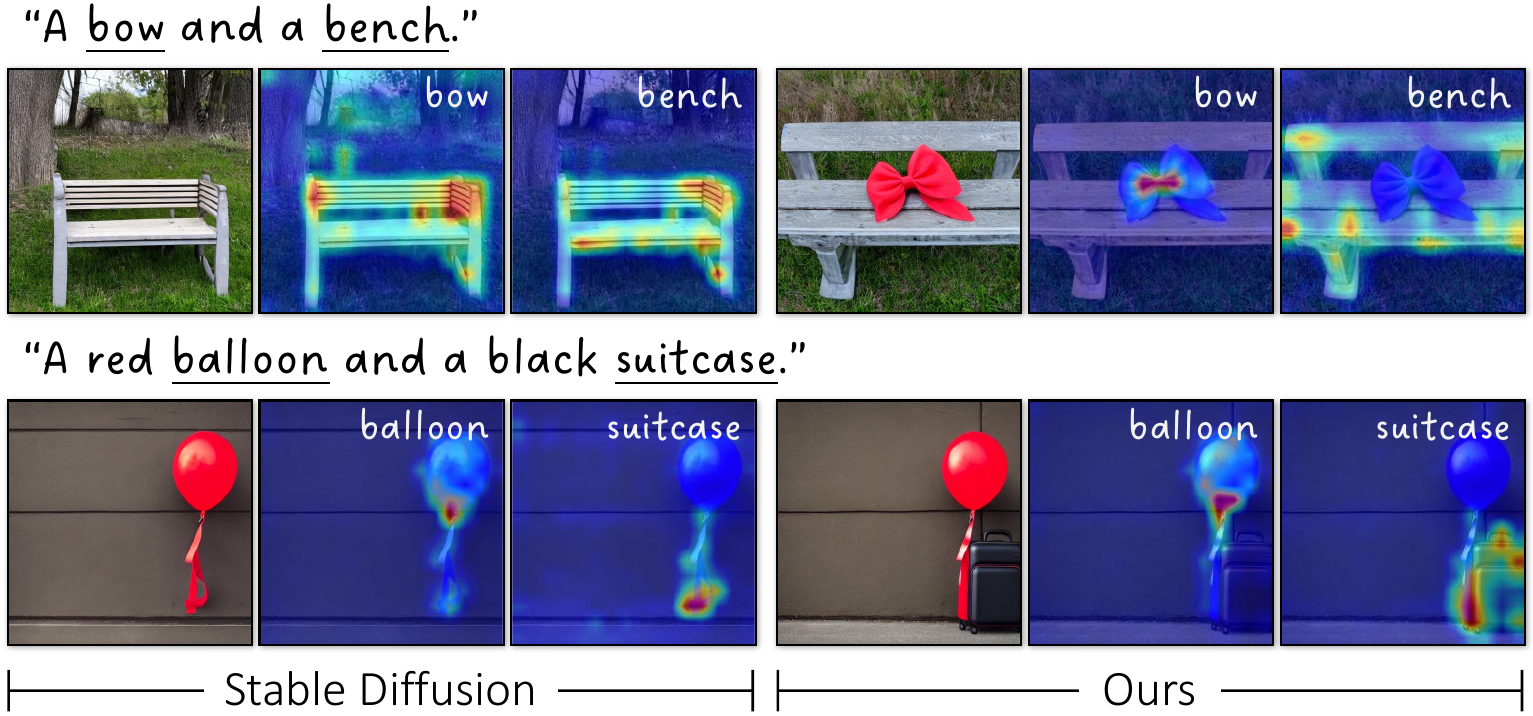}
\caption{\textbf{Visualization of the final cross-attention maps.}}
\label{fig:ae-cross-attention}
\end{figure}

\noindent \textbf{Visualization of the attention maps.} Following \cite{chefer2023attend}, we also provide a visualization of the final cross-attention map for each subject token after denoising process in Fig.~\ref{fig:ae-cross-attention}. Our method achieves reasonable attention allocation, leading to semantically-consistent results.

\section{Conclusion}
\label{sec:conclusion}

The core of our methed is the Initial Noise Optimization (\textsc{InitNO}), which consists of the initial latent space partitioning and the noise optimization pipeline, responsible for defining valid regions and steering noise navigation, respectively. Specifically, we investigate the utility of attention maps in the diffusion model, harnessing them to formulate the cross-attention response score and the self-attention conflict score, both instrumental to the partitioning of the initial latent space. Moreover, we bypass the complex balance between under-optimization and over-optimization of noise via a meticulously crafted pipeline that introduces a novel distribution alignment loss. We further demonstrate the superiority of our approach over existing state-of-the-art methods in generating semantically-faithful images, providing a versatile plug-and-play solution that effectively integrates into existing diffusion models for training-free controllable generation

\section*{Acknowledgment} This work is partly supported by the National Key R\&D Program of China (2022ZD0161902), the National Natural Science Foundation of China (62022011, 62202031), the Research Program of State Key Laboratory of Software Development Environment (SKLSDE-2023ZX-14), and the Fundamental Research Funds for the Central Universities. We also give specical thanks to Alibaba Group for their contribution to this paper.
\clearpage
{
    \small
    \bibliographystyle{ieeenat_fullname}
    \bibliography{main}
}

\end{document}